# A novel action recognition system for smart monitoring of elderly people using Action Pattern Image and Series CNN with transfer learning

L. Aneesh Euprazia[a] and K.K.Thyagharajan[b,*]

[a] *anueuprazia@gmail.com, Department of Computer Science and Engineering, RMD Engineering College, Chennai, India;*

[b] *Department of Electronics and* Communication *Engineering, RMD Engineering College, Chennai, India*

*\*corresponding author, e-mail: kkthyagharajan@yahoo.com, RMD Engineering College, Kavaraipettai, Gummidipoondi Taluk, Thiruvallur Dist., Tamil Nadu, INDIA. Pin. 601206*

**Abstract:** Falling of elderly people who are staying alone at home leads to health risks. If they are not attended immediately even it may lead to fatal danger to their life. In this paper a novel computer vision-based system for smart monitoring of elderly people using Series Convolutional Neural Network (SCNN) with transfer learning is proposed. When CNN is trained by the frames of the videos directly, it learns from all pixels including the background pixels. Generally, the background in a video does not contribute anything in identifying the action and actually it will mislead the action classification. So, we propose a novel action recognition system and our contributions are 1) to generate more general action patterns which are not affected by illumination and background variations of the video sequences and eliminate the obligation of image augmentation in CNN training 2) to design SCNN architecture and enhance the feature extraction process to learn large amount of data, 3) to present the patterns learnt by the neurons in the layers and analyze how these neurons capture the action when the input pattern is passing through these neurons, and 4) to extend the capability of the trained SCNN for recognizing fall actions using transfer learning.

# 1. Introduction

Nowadays staying elderly persons at home alone is unavoidable. In order to satisfy the basic needs and to keep up our life style in the society, both husband and wife are pushed to earn money. If older people with health complaints are staying alone at home then the family members who are at work cannot concentrate in their work and they cannot be in peace of mind. Meanwhile the older people are also prone to health risks. To avoid such pathetic situations smart monitoring of elderly people was introduced and more researchers [1] [2] [3] are giving attention to that. By using such systems the family members or caretakers can get to know whether the elderly persons at home are safe or not and if any threat action happens to them immediate attention can be provided to rescue them.

Various types of fall detection systems have been developed to support the elderly people. Most of those systems use wearable sensors. The commonly used sensors in fall detection systems are motion sensor, depth sensor and thermal sensor. The thermal sensor-based systems use heat of the subject as the basic component. Each pixel value of sensor data acts as a source and helps in detecting the actions. Motion sensors use the key features like acceleration, velocity and displacement to detect actions. From these features the tri-axial body acceleration, magnitude of body acceleration and angular velocity are computed. These systems need a wearable device to be worn by the subject under investigation in any part of the body. In RFID (Radio-frequency identification) tag-based fall detection system a RFID tag has to be worn by the subject [4].
Initially wireless beacon signals are propagated and the tags will reply to that. The Received Signal Strength and Doppler frequency values are used to detect actions. The systems which are relying on sensors implicitly insist the subjects to wear devices on their body. Mostly elderly people won't feel comfortable to wear a device or they may forget to wear device. These are the problems with wearable device based fall detection system.

Radar based fall detection system [5] sends continuous wave on the subject and collects the signals backscattered from human subject. Those signals are represented as range map and spectrogram which are then processed to detect human actions. But continuous propagation of wave on subject will not be good always. So a system without using wearable devices and signal propagation is advisable for monitoring elderly people. Vision based monitoring system is free of wearable devices and frequent propagation of signals. These systems need a camera to record the human actions continuously and when a threat action is happening it will intimate to responsible persons. The CNN [6] [7], LSTM (Long short-term memory) [8], SVM (Support Vector Machine) [9] [10] can be used to detect actions. In this work we developed computer vision-based action detection system with fall detection by proposing a novel Series CNN architecture. Inspired by automated feature extraction, high recognition rate and simple architecture for a large dataset the Series CNN is opted in this work. The novelties of this work are listed as follows: 1) to generate more general action pattern inputs for CNN from the video sequences by eliminating the obligation of image augmentation. These patterns are called Action Pattern Images (API) and they are also independent from illumination variations and the background variations 2) to design a Series CNN (SCNN) architecture to enhance and simplify the feature extraction process of CNN for learning large amount of data, 3) to present the patterns produced by the weights of the neurons in the layers and analyze how these neurons capture the actions when the input pattern is passing through these neurons, 4) to extend the recognition capability of the trained SCNN by creating a new network and transferring the knowledge learnt by SCNN to it and then train the new network with additional new data of fall action.

The remaining part of this paper is organized as follows: Section 2 summarizes the research work already carried out in this field of study. Section 3 provides the description of the datasets used in this work. Section 4 explains the proposed system. Section 5 includes the experimental results and analysis and section 6 provides the conclusion.

## 2. Literature survey

Chua et al. [1] extracted the foreground subject using background subtraction. The bounding box of foreground subject is computed and it is splitted into three regions. The height, width and centroid of each region are computed. Using the three centroid points two lines were traced and length of those lines were calculated. The angle between each line and the horizontal x axis was computed. Since this angle varies with respect to action, it is used as a basic feature to discriminate actions. Harrou et al. [2] introduced the concept of splitting the human silhouette into five parts by drawing lines from centroid and fixing angles (Figure 1). But for a very short and stout person the A3, A4 and A5 regions remain almost same for fall and non-fall actions which may affect the fall detection rates.

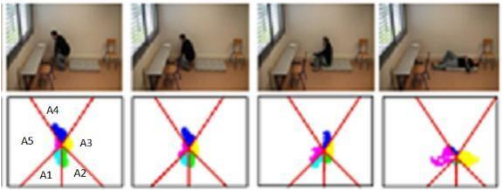

Fig.1. Partitioning human body into five sections

Rougier et al. [3] proposed shape based detection of fall by obtaining movable edge points. Mean matching cost and full Procrustes distance were computed and fed into Gaussian Mixture Model (GMM) to detect fall and non-fall activities. Our work uses edges as the basic component with less computation complexity compared to the work done in Rougier et al. [24] which uses log polar histogram and estimates the matching cost on edge points with more computation complexity. The fall action detection system developed by Zhu et al. [4] uses Received Signal Strength (RSS) and Doppler frequency. Their system relies on usage of RFID tag around object's neck. This system sends fall message to recipient after particular (alarm) period of time. Hence after the fall a person at risk state may retain consciousness and if that person is not interested to intimate this information to the recipient, during the alarm period that person can stop sending fall message to recipient and it is a drawback of that system. Hsieh and Jeng [6] used optical flow features and action rules are provided to feedback optical flow CNN to discriminate threat events and safe events. Nunez-Marcos et al. [7] used TVL-1 optical flow algorithm to generate optical flow images of video sequences. The obtained optical flow images are fed into CNN to detect fall or non-fall activity. Our work also uses CNN for action recognition but does not use optical flow images. Lu et al. [8] extracted the feature cubes from video sequence for each 16 frames group by a 3D CNN. The action is predicted by LSTM. Hussain et al. [9] proposed a fall detection system using wearable sensors. For every 10 seconds the sensed data is acquired. Acceleration and angular rotation acts as the key points to detect falls. The classification performance is analysed with SVM, Random forest and KNN (k-nearest neighbour) classifiers. Saleh and Jeannes [10] constructed a 12-component feature vector from the acceleration signals which is then given to SVM for classification. Chelli and Patzold [11] proposed a fall detection system that relies on a smartphone equipped with gyroscope. Through the acceleration and angular velocity data given by mobile phone the time domain features and frequency domain features are generated. But midrange mobile phones are not equipped with gyroscope, so all mobile phone users cannot use this system. The mobile phone may also break due to mishandling, and they need regular charging. Euprazia and Thyagharajan [12] traced the minimum bounding box of moving object to obtain the features such as major axis length, minor axis length and orientation angle. These features are used discriminate the daily living activities and fall activity. Georgakopoulos et al. [13] constructed binary silhouettes and used it as CNN input. Features extracted by CNN is fused with Zernike moments of binary silhouttes and processed by feed-forward neural network (FNN) for classification. Motion sensors such as passive infrared sensors (PIR) and door sensors were used by Gochoo et al. [14] to monitor elderly people. The 2D representation of those sensor data is taken as the activity image and it is given to deep CNN to identify the action. Jing et al. [15] proposed a method for generating proxy video of fixed length from input video sequence for action recognition. Through even sampling and random sampling, the proxy videos were generated. To handle proxy videos of varying length eight networks were created. Our work is also using a newly created network but the same single network is used to learn all datasets and also to learn any new dataset using transfer learning. Kepski and Kwolek [16] proposed fall detection through combined result of acceleration signals and depth data. By fixing threshold for total sum vector obtained from acceleration of x, y, z axis and depth features the fall actions are detected. Kwon et al. [17] introduced TSAM (Temporal Signal Angle Measurement) algorithm to classify various fall types using signals. The fall types are determined using the angle between the test signals and reference signals. Liu et al. [18] introduced the concept of reduction of sampling rate of sensed data. His work mainly focused on lowering the sensor power consumption and increasing the processing time of fall detection system. Shape, head position and motion information-based fall detection was proposed by Lotfi et al. [19]. The moving object was approximated by ellipse and the ellipse was approximated by rectangle. The extracted features were given to neural network for classification. Luo et al. [20] used depth sensor and thermal sensor to preserve privacy of subject who is under activity monitoring. The long video sequence was segmented into clips with T frames and the softmax scores of clips were averaged to obtain the frame level scores. Along with that the depth and thermal features were combined, and then the softmax scores were obtained to make final prediction. Min et al. [21] used Faster R-CNN (Region-based CNN) to obtain object's location in a scene. After locating the object the features such as human shape aspect ratio, centroid and motion speed are obtained to detect fall and other activities of daily living (ADL) activities. Quadros et al. [22] proposed wearable device based fall detection system by using acceleration, velocity and displacement as basic components. Taramasco et al. [23] used thermal sensor to monitor fall action by measuring the heat received in two planes inside the room. First plane is set as 10 cm above the floor and second plane is 100 cm above the floor. Since the body temperature of the human subject will not be constant always, it may affect the fall detection. Yu et al. [24] used acceleration signals to detect fall activity. The acceleration signals obtained from sensors are transformed into a vector. By performing computations on vector, vertical and horizontal components are obtained and they are fed into Hidden Markov

Model (HMM) to detect falls. Zhao et al. [25] created cascade CNN-HsNet with three networks S1, S2 and S3 to detect pedestrian in short term events. The S1 network will pull down negative samples, network S2 reduce false detection and S3 network is responsible for improvising learning and thereby increases the discriminative power. Since these networks are cascaded, the non-performance of one network will affect other network. Zheng et al. [26] used primitive edges chosen from set of localized edges of a video sequence. The VGG19 net is used to extract features with different pooling methods. Our work also relies on edges of the foreground object but all the edges of the video sequence are considered for better recognition.

## 3. Datasets

The bench mark datasets used for evaluating action recognition systems are discussed in this section.

### 3.1 KTH dataset [31]

The KTH dataset has 6 action classes such as boxing, handclapping, hand waving, jogging, running and walking. The actions are performed by 25 subjects in four different scenarios and recorded with a static camera. Handclap action class has 99 video sequences and rest of the action classes have 100 video sequences each. So, this publicly available dataset has 599 video sequences. The sample frames taken from KTH dataset is given in Figure 2.

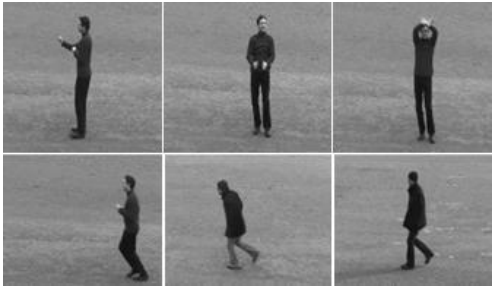

Fig.2. Sample frames of KTH dataset

## 3.2 Multiple cameras fall dataset [28]

The Multiple cameras fall dataset has video sequences of different fall actions recorded with 8 IP video cameras. Each fall action is performed by single subject and it is recorded by 8 cameras, those 8 video sequences are considered as a scenario. This dataset has 24 scenarios with 192 video sequences. The sample frames taken from this dataset is given in Figure 3.

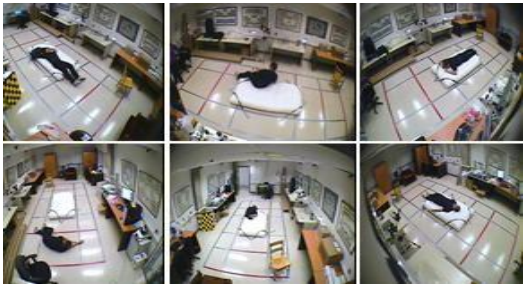

Fig.3. Sample frames taken from Multiple cameras fall dataset.

## 3.3 Fall detection dataset [29]

The Fall Detection dataset contains 191 video sequences of actions classes falling, walking, mobbing, sitting and office activities. The video sequences are recorded in four different locations: Home, Coffee room, Office and Lecture room. Figure 4 shows the sample frames of Fall Detection dataset.

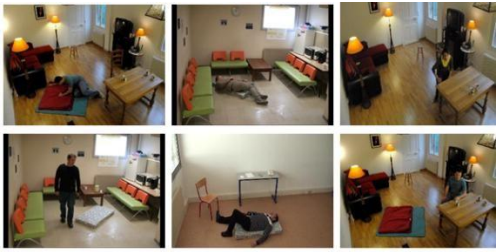

Fig.4. Sample frames of Fall Detection dataset

### 3.4 SisFall dataset [30]

The SisFall dataset has 19 video sequences of activities of daily living(ADL) (walking slowly, jogging, walking upstairs, slowly sit, standing and slowly bending, gentle jump…) and 15 fall videos ( forward fall, backward fall, lateral fall and vertical fall). So, on the whole it is having 34 video sequences. The sample frames of SisFall dataset is shown in Figure 5.

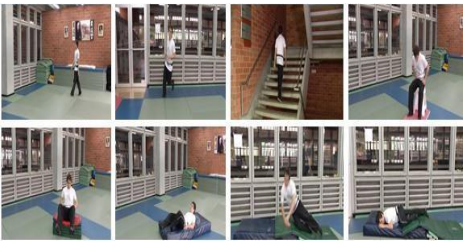

Fig.5. Sample frames of SisFall dataset

### 3.5 UR fall detection dataset [32]

The UR fall detection dataset has 71 video sequences of fall activity and activities of daily living. It has 40 ADL videos (sitting on chair, lying on bed, searching objects on floor). It has 31 fall video sequences (fall from chair, forward fall.). The sample frames of UR fall detection dataset is shown in Figure 6.

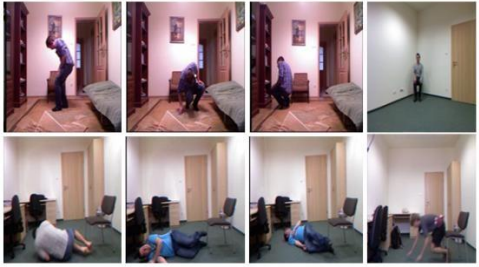

Fig.6. Sample frames of UR fall detection dataset

## 4. Proposed work

The block diagram of the proposed system is shown in figure 7. This system has been designed to classify five actions which are used to monitor healthy elderly people. The preprocessing block converts the action video into action pattern image (API). This API gets the background in the video removed and it is also independent from illumination variation. This API is given as input to SCNN. The features extracted by SCNN layers are classified to identify the action performed. The second block diagram shown in figure 8 depicts how to convert the existing SCNN to train and recognize a new action called fall action. Here we transfer the knowledge learnt in recognizing the actions such as walking, hand wave, hand clap, jogging and running by the previous SCNN to the new CNN by transferring the trained SCNN layers. We remove the classifier for five actions of SCNN and add a new classifier to recognize six actions as shown in figure 8.

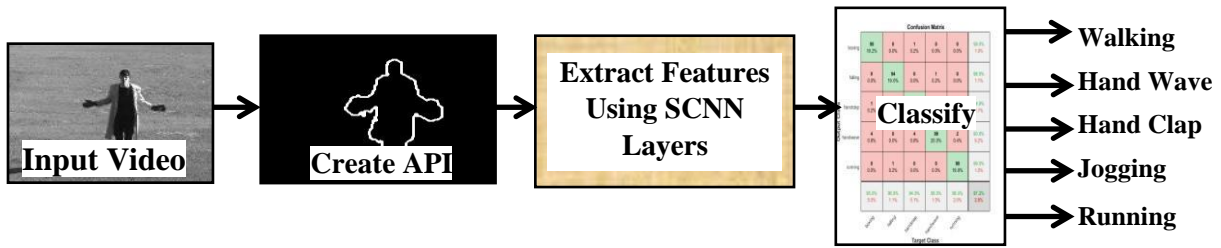

Fig.7. Recognizing five general actions

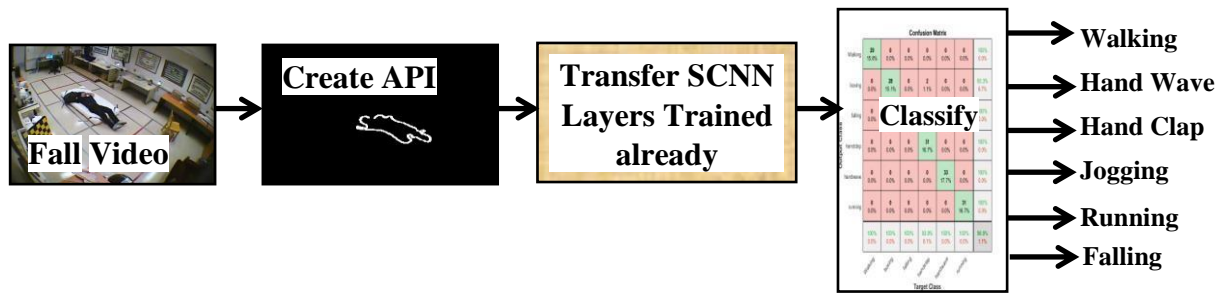

Fig.8. Transfer learning with additional falling action

To train and test the SCNN, we collect the required videos from the bench mark data bases discussed in section 3. The videos available for different actions are summarized in table 1. Since we are not able get all the six required actions from a single data base, we created a data base by combining these benchmark data bases. For example, in the Fall Detection data base 'walking' action is accompanied with 'sitting' and 'fall' actions and there are no separate videos for walking action. In the Sisfall data base 'standing' action is accompanied with sitting action and no separate video available for standing action. Table 2 shows the number of videos we used for six actions to train and test our system. Multiple cameras fall data base has video sequences of fall action at different angles. Experimenting homogeneous data will give good result but cannot give assurance of accurate detection when it comes to heterogeneous environment. So for enabling heterogeneity the KTH dataset and the Multiple camera fall dataset are combined. The video sequences of KTH dataset and video sequences of scenarios 1,2,3,4,5,6,7,8,11,12,13,14,15 (100 video sequences) of Multiple cameras fall dataset are chosen for training and testing our system.

Table 1: Summary of videos available in various databases

| Action | Number of videos in the databases | | | | | Total |
|---|---|---|---|---|---|---|
| | KTH | Multiple Camera | Fall Detection | SisFall | UR fall | |
| Sitting | NIL | NIL | 33 | 7 | 9 | 49 |
| Bending | NIL | NIL | NIL | 2 | NIL | 2 |
| Falling | NIL | 192 | 70 | NIL | 31 | 293 |
| Lying | NIL | NIL | NIL | NIL | 11 | 11 |
| Standing | NIL | NIL | NIL | No separate videos | NIL | 0 |
| Boxing | 100 | NIL | NIL | NIL | NIL | 100 |
| Handclap | 99 | NIL | NIL | NIL | NIL | 99 |
| Handwave | 100 | NIL | NIL | NIL | NIL | 100 |
| Running | 100 | NIL | NIL | NIL | NIL | 100 |
| Jogging | 100 | NIL | NIL | 2 | NIL | 102 |
| Walking | 100 | NIL | No separate videos | 2 | NIL | 102 |

Table 2: Number of videos used for Training and Testing

| Actions | No. of videos | Video Data bases used |
|---|---|---|
| Handclap | 99 | KTH [31] |
| Handwave | 100 | KTH [31] |
| Jogging | 102 | KTH [31], SisFall [30] |
| Running | 100 | KTH [31] |
| Walking | 102 | KTH [31], SisFall [30] |
| Falling | 150 | Multiple Camera [28], FDD [29], SisFall [30], URFD [32] |

### 4.1 Action Pattern Image (API)

This section proposes a novel method to generate action pattern images (API) from the action video sequences which are given as inputs to the CNN. These APIs eliminate the necessity of creating augmented images while training the CNN. These are also independent from illumination variations and the background variations. Use of API reduces the training time. Generally CNN learns from the pixels of the video frames. When these frames are directly fed to CNN, the learning will be biased based on background pixels also. A large memory will also be required to store these frames. Similar video with different background will be learnt by the network as a different one i.e. the generalization capability will be less. But the proposed pre-processing reduces the amount of data to be fed to CNN without losing the recognition capability and increasing the generalization capability with reduced computational requirements. To create these action patterns first we have to extract the foreground information from a video sequence. We have to either remove or suppress the background. To remove simple and complex background both background subtraction and intensity adjustments are done together. The proposed network is designed to recognize images with resolution of 256 x 256 pixels input. Therefore, each frame of the video is resized to 256 x 256 pixels. First, the background in each frame is removed by background subtraction method and then the foreground pixels in a frame are enhanced if the normalized intensity value of a pixel is greater than 0.3. From the enhanced foreground data, the edge features are extracted. The edges of human object present in each frame are extracted using 'Canny' edge detectors and accumulated together resulting Action Pattern Image (API). The API extraction process is implemented in Matlab 2018a environment and the algorithm is given below. The pictorial representation of the above process is given in Figure 9.

Algorithm:

Read video sequence

N = Number of frames

Accumulate Edges = Null Image

For i = 1 : 2 : N

    Read frame[i]

    Perform background subtraction on frame[i]

    Normalize the pixel values

    Enhance the pixel if intensity level is greater than 0.3

    Img = convert RGB to GRAY (frame[i])

    Edge = edge ( Img, 'Canny' , 'vertical')

   Accumulate Edges = Accumulate Edges + Edge

End

Action Pattern Image = Outline of Accumulate Edges

The APIs of actions such as boxing, handclap, hand wave, jogging, running and falling are given in Figure 10. The size of the API generated from each video clip is 256x256x1 and this API is fed into the proposed Series CNN as input. The design and architecture of the prosed CNN is given in the next section.

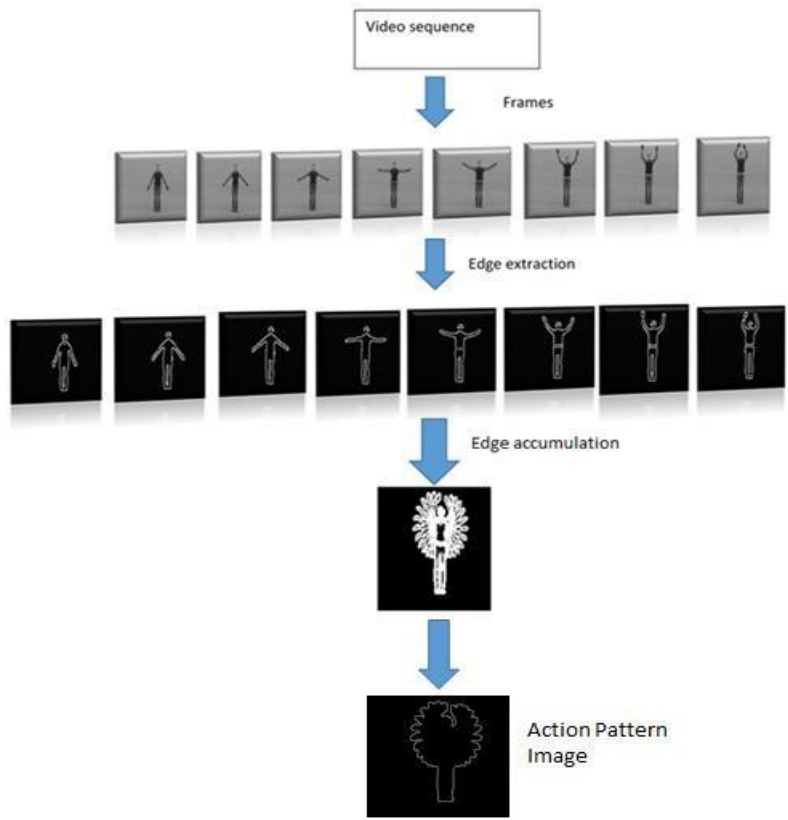

Fig.9. Creating API from a video sequence

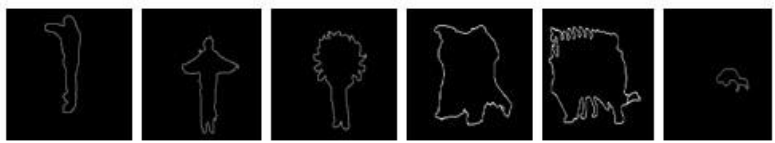

Fig. 10 Action Pattern Images (APIs) for different actions

Table 3: APIs created for Training and testing the SCNN

| Actions | No. of APIs | Video Data bases used |
|---|---|---|
| Handclap | 100 | KTH [31] |
| Hand wave | 100 | KTH [31] |
| Jogging | 100 | KTH [31], SisFall [30] |
| Running | 100 | KTH [31] |
| Walking | 100 | KTH [31], SisFall [30] |
| Falling | 150 | Multiple Camera [28], FDD [29], SisFall [30], URFD [32] |

To train the proposed CNN architecture we created a dataset of Action Pattern Images (APIs) for six actions. Ninety-nine APIs were created from 99 handclap videos of KTH data base and one API from these APIs was selected randomly and added with this dataset to get 100 APIs for handclap dataset. The APIs for jogging and walking were created from the videos obtained from KTH and SisFall databases. We randomly selected 100 APIs out of 102 APIs created from 102 videos obtained for each of these two categories. These are summarized in table 3. We trained the Series CNN for the actions given in table-3, and then we used transfer learning method to train falling action. The APIs for falling actions were created from randomly selected 150 videos obtained from Multiple Camera, FDD and SisFall data bases. Out of 150 APIs only 100 were randomly selected for training the transfer learning network. Similarly portions of (20%) API datasets for other actions were also included for training the transfer-learning network.

## 4.2 CNN architecture

The architecture of the proposed CNN is illustrated in this section. In this architecture the following layers are used: A Input layer of size 256x256x1 with zero center normalization, fourteen convolution layers (CX) with 14 batch normalization (BX) and 14 ReLU (Rectified Linear Unit) layers where 'X' (1 to 14) is the layer number. Batch Normalization layers are used to store the normalized values got from the outputs of the previous layers and to give as inputs to the next upper level layers. Non-linearity is introduced in the output of the layer by using ReLU activation layers. We use max pooling layers with first 6 ReLU layers and $11^{th}$ & $14^{th}$ ReLU layers because if pooling is used after each ReLU then we may not increase the depth of the network and the number of parameters used. Using a smaller number of layers and parameters will overfit the data on the network. We use two fully connected (FC) layers. The second FC is followed by a softmax layer and a classification layer as shown in figure 11. The convolution layer 1 (C1) uses 8 kernels of size 3x3 and it produces 8 channels, C2 has 16 kernels of size 3x3, C3 uses 32 kernels of size 3x3 and the convolution layers from C4 to C14 use kernels as given in table 4. All the convolution and Max pooling layers are extended by one padding bit on all sides. Max pooling is performed with kernel size of 2x2. Since the layer size after introducing Max pooling-7 layer is very small we include convolution layers 12 and 13 with padding but without pooling to increase the depth of the network. We include a Max pooling layer after convolution layer 14 with padding. Two FCs are used with 2048 and 5 neurons respectively. Since the batch normalization layers and ReLU layers do not decide the number of parameters used, the network layer details is summarized in table 4 without those layers but included in the pictorial representation of the architecture shown figure 11.

Table 4: Details of CNN Layers of the proposed architecture

| S. No. | Layer type | Layer Size | Kernel size (wxh) | stride | padding | No. of kernels / channels (c) | Output Size from the Layer | No. of parameters |
|---|---|---|---|---|---|---|---|---|
| 1 | Input | 256 x256x3 | - | - | - | - | - | - |
| 2 | Convolution-1 (C1) | 256 x256 | 3 x 3 | 1 | 1 | 8 | 256 x 256 | 80 |
| 3 | Max Pooling-1 | 256 x256 | 2 x 2 | 2 | 0 | 8 | 128 x 128 | - |
| 4 | Convolution-2 (C2) | 128 x128 | 3 x 3 | 1 | 1 | 16 | 128 x 128 | 160 |
| 5 | Max Pooling-2 | 128 x128 | 2 x 2 | 2 | 0 | 16 | 64 x 64 | 0 |
| 6 | Convolution-3 (C3) | 64x64 | 3 x 3 | 1 | 1 | 32 | 64 x 64 | 320 |
| 7 | Max Pooling-3 | 64x64 | 2 x 2 | 2 | 0 | 32 | 32x32 | - |
| 8 | Convolution-4 (C4) | 32x32 | 3 x 3 | 1 | 1 | 64 | 32x32 | 640 |
| 9 | Max Pooling-4 | 32x32 | 2 x 2 | 2 | 0 | 64 | 16x16 | - |
| 10 | Convolution-5 (C5) | 16x16 | 3 x 3 | 1 | 1 | 128 | 16x16 | 1280 |
| 11 | Max Pooling-5 | 16x16 | 2 x 2 | 2 | 0 | 128 | 8X8 | - |
| 12 | Convolution-6 (C6) | 8X8 | 3 x 3 | 1 | 1 | 256 | 8X8 | 2560 |
| 13 | Max Pooling-6 | 8X8 | 2 x 2 | 2 | 0 | 256 | 4X4 | - |
| 14 | Convolution-7 (C7) | 4X4 | 3 x 3 | 1 | 1 | 512 | 4X4 | 5120 |

| 15 | Convolution-8 (C8) | 4X4 | 3 x 3 | 1 | 1 | 1024 | 4X4 | 10240 |
| --- | --- | --- | --- | --- | --- | --- | --- | --- |
| 16 | Convolution-9 (C9) | 4X4 | 3 x 3 | 1 | 1 | 1024 | 4X4 | 10240 |
| 17 | Convolution-10 (C10) | 4X4 | 3 x 3 | 1 | 1 | 1024 | 4X4 | 10240 |
| 18 | Convolution-11 (C11) | 4X4 | 3 x 3 | 1 | 1 | 1024 | 4X4 | 10240 |
| 19 | Max Pooling-7 | 4X4 | 2 x 2 | 2 | 0 | 1024 | 2X2 | - |
| 20 | Convolution-12 (C12) | 2X2 | 3 x 3 | 1 | 1 | 2048 | 2X2 | 20480 |
| 21 | Convolution-13 (C13) | 2X2 | 3 x 3 | 1 | 1 | 2048 | 2X2 | 20480 |
| 22 | Convolution-14 (C14) | 2X2 | 3 x 3 | 1 | 2 | 2048 | 2X2 | 20480 |
| 23 | Max Pooling-8 | 2X2 | 2 x 2 | 2 | 0 | 2048 | 2X2 | - |
| 24 | Fully Connected-1 | 2048 | - | - | - | - | - | 8192 |
| 25 | Fully Connected-2 | 2048 | - | - | - | - | - | 4194304 |
| Total number of parameters | | | | | | | | 4,315,056 |

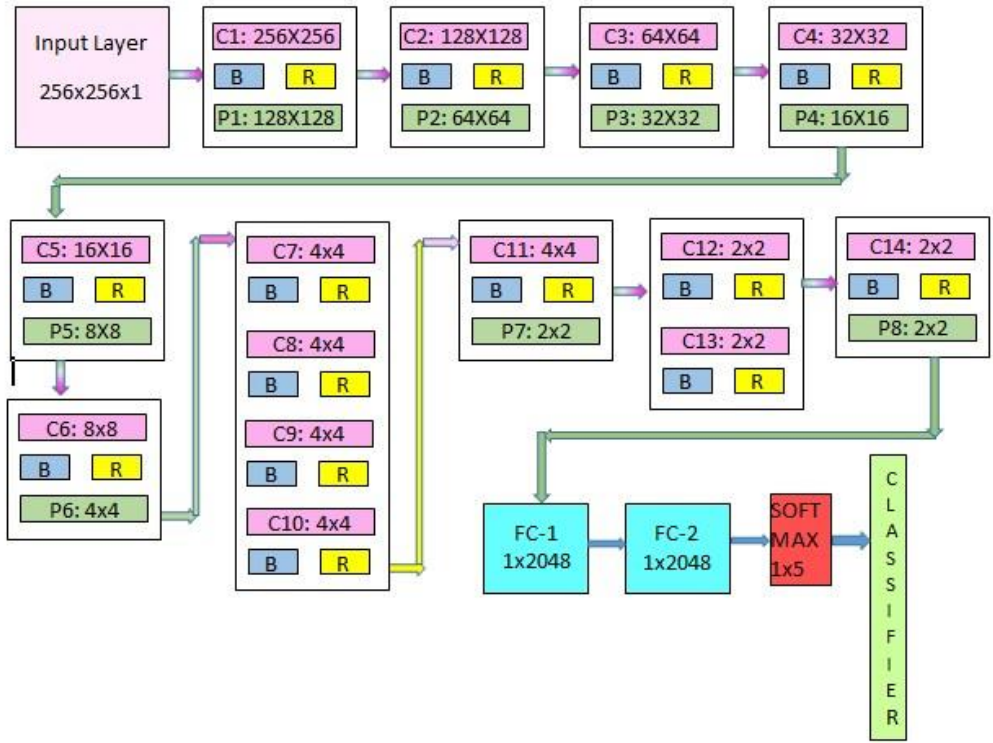

Fig.11. Architecture of CNN

## 5. Experimental results and analysis

Our system was implemented using MATLAB 2018a on a PC having Intel(R)) Core (TM) i5-2410M CPU @ 2.30 GHz with 4GB memory. The training and testing was done on single CPU.

### 5.1 Experiment 1

The network was created and set with the following training options: Stochastic Gradient Descent with Momentum (SGDM) optimizer, the learning rate schedule is piecewise, learning rate of 0.001 with momentum 0.9, L2 Regularization 0.004, mini-batch size 45 and learn rate drop factor of 0.1 per 8 epochs. We assigned 10 iterations per epoch. This architecture uses 4,315,056 parameters as shown in table 4. We set the mini-batch size to 45, and randomly selected 90 APIs from each category of action as given in table 5 for giving as training input to SCNN. The maximum number of epochs for training the network was set to 30 and the training progress was observed for accuracy and mini-batch loss as shown in figure 12. It was noticed that the accuracy reached 100% and mini-batch loss became low after 150th iteration. Table 6 shows that the loss decreases for further iterations. Since it is very slowly decreasing and no improvement in accuracy is possible we allowed training for few more iterations for further stabilization of error and stopped the training at 188th iteration.

Table 5: APIs used for Training

| Actions | API | Video Data bases used |
|---|---|---|
| Handclap | 90 | KTH [31] |
| Hand wave | 90 | KTH [31] |
| Jogging | 90 | KTH [31], SisFall [30] |
| Running | 90 | KTH [31] |
| Walking | 90 | KTH [31], SisFall [30] |

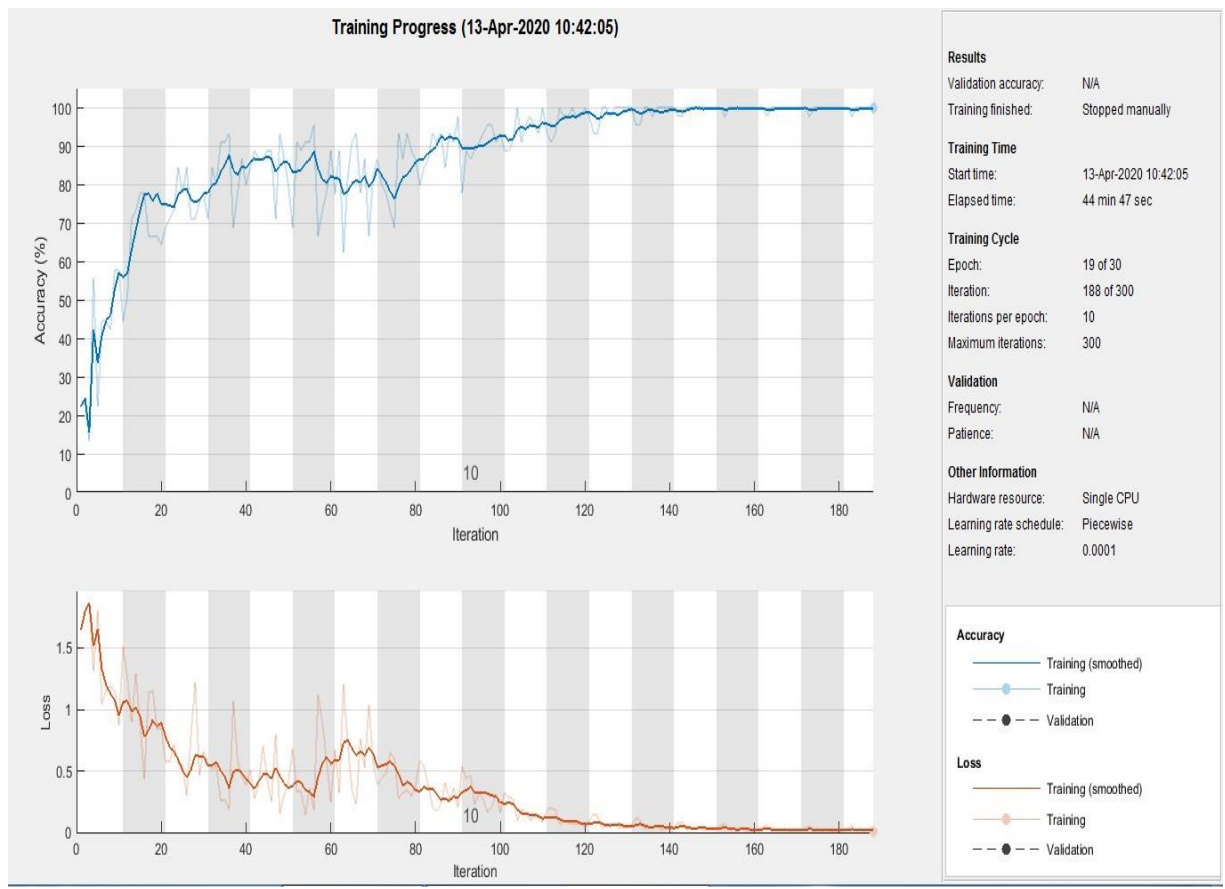

Fig. 12. Training Progress with mini-batch size of 45

Table 6: Mini-batch loss and accuracy with mini-batch size of 45

| Epoch | Iteration | Time Elapsed (hh:mm:ss) | Mini-batch Accuracy | Mini-batch Loss | Base Learning Rate |
|---|---|---|---|---|---|
| 1 | 1 | 00:00:36 | 22.22% | 1.6456 | 0.0100 |
| 5 | 50 | 00:14:01 | 80.00% | 0.4112 | 0.0100 |
| 10 | 100 | 00:25:23 | 93.33% | 0.1637 | 0.0010 |
| 15 | 150 | 00:36:24 | 100.00% | 0.0197 | 0.0010 |
| 19 | 188 | 00:44:47 | 100.00% | 0.0114 | 0.00010 |

After training was completed we tested the trained SCNN with the all 100 APIs given in table 3 and the confusion matrix obtained in this test is shown in figure 13. The confusion matrix shows that the actions are recognized with 96.8% over all accuracy. The classification accuracy improved to 97.6% when we trained the SCNN with mini-batch size of 50 and allowed up to 232 iterations But this increases the training time to 56 minutes and 29 seconds instead of the previous training done in 44 minutes and 47 seconds. It is also observed that slight increase or decrease in accuracy may also happen due to the random selection of APIs for training. However, this system exhibits good accuracy (more than 94%) for classification of all five action videos.

Fig. 13. Confusion matrix of test 1

From the figure 13 it is observed that out of 500 APIs only 16 were misclassified. Figure 14 (a), (b) and (c) gives three the hand wave actions misclassified as hand clapp action and figure 14d shows the one hand wave action misclassified as jogging action. Similarly two walking actions have been misclassified as jogging action and one hand clap action has been misclassified hand wave action and so on.

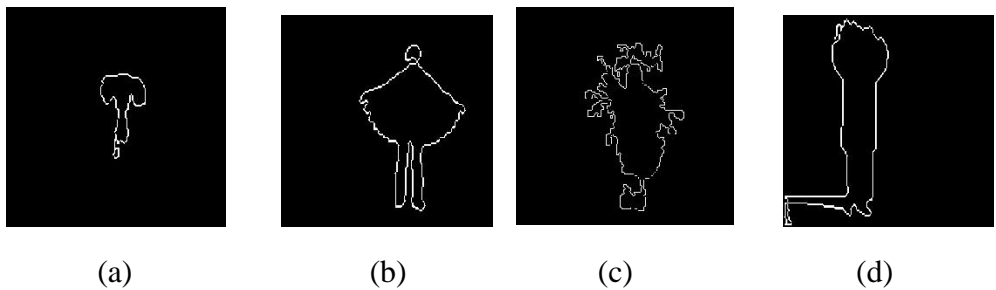

(a)　　　　　　　(b)　　　　　　　(c)　　　　　　　(d)

Fig.14. Hand wave actions misclassified as hand clap and jogging actions

When API for hand clap shown in figure 15 was given as input, the features were captured by convolution layer 2 as shown in figure 16 (a) and the features were captured by convolution layer 6 as shown in figure 16 (b). It is observed that deeper layers contain higher number of features constructed using lower level of features from earlier layers. The weight pattern that was stored for convolution layer during training is shown in figure 17. When input API passes through this weight pattern the neurons in the convolution layer capture the features of input. Since, the input API is binary image, the complete shape and coarse details are captured at this level. At convolution layer 6, minute details are captured because more number of filters are used at this level.

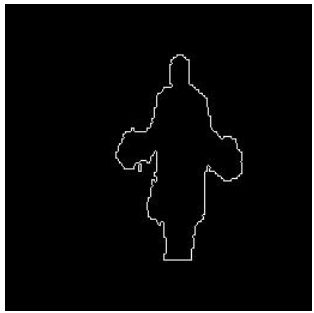

Fig.15. Input Pattern

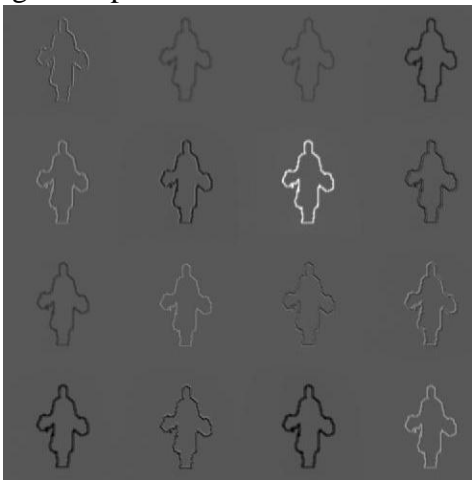 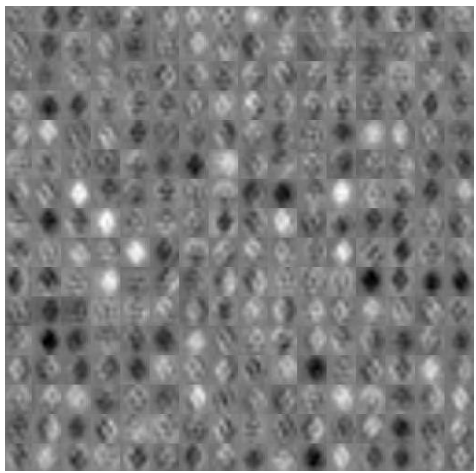

(a) 16 channels of Convolution layer 2     (b) 256 channels of Convolution layer 6 Fig.16. Features Captured by each channel in Convolution Layers

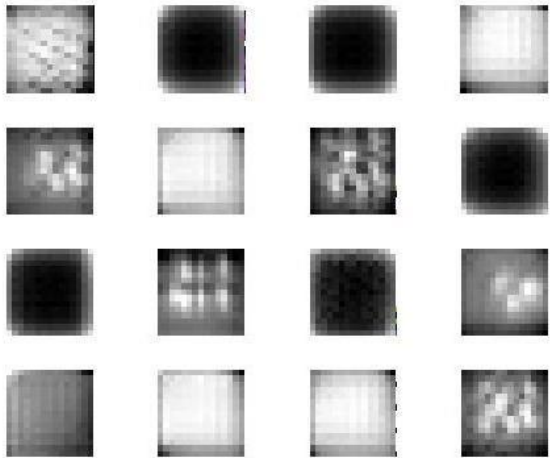

Fig.17. Weight pattern of convolution layer 2

### 5.2 Experiment 2

When new classes of actions are to be recognized by the SCNN already trained for other actions, it may require creating a new network and train with all data in all classes because number of classes to be recognized by the new CNN has increased. To avoid training with complete set of data again and reduce the training time, we propose to create a new CNN and transfer the knowledge from the already trained SCNN. We removed the softmax and classification layers of old SCNN after transferring the layers to the new network and added new sofmax and classification layers to support six classes of actions including fall action. Then we train the new network for fall action with all APIs in the new data set and a small portion of the old data for other actions. In this experiment, we trained 100 APIs of a new class 'fall action' along with 20% APIs of old classes. The videos for fall action are randomly chosen from three different data bases as shown in table 7. The same values of training parameters used by the old network were applied without any change because it was observed that increasing the learning rate made the new network not to recognize some of the APIs already trained even though the training time for the new network is reduced. Figure 18 shows the progress of training in the transfer learning. It is observed that the accuracy reaches 100% at $30^{th}$ iterations and remains constant after that. The mini-batch loss is also not decreasing much after $30^{th}$ iteration. So, the training was stopped at $39^{th}$ iteration and the time required for training is 10 minutes and 47 seconds as given in table 8.  Table 7: APIs used for Transfer Learning

| Actions | No. of APIs | Video Data bases used |
|---|---|---|
| Handclap | 20 | KTH [31] |
| Hand wave | 20 | KTH [31] |

| Jogging | 20 | KTH [31], SisFall [30] |
| Running | 20 | KTH [31] |
| Walking | 20 | KTH [31], SisFall [30] |
| Falling | 100 | Multiple Camera [28], FDD [29], SisFall [30], URFD [32] |

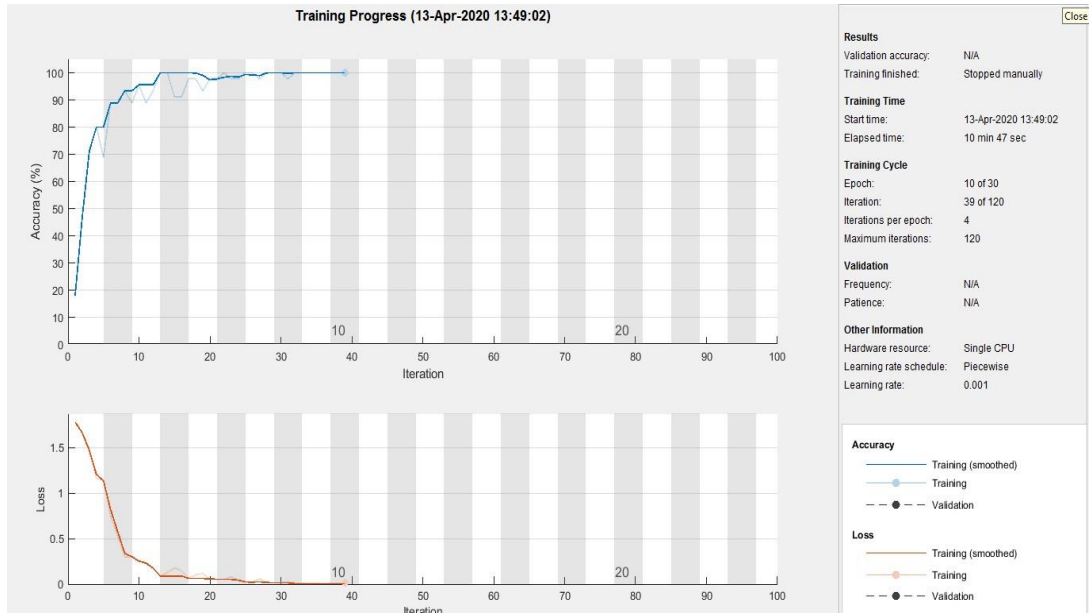

Fig.18. Transfer learning Training Progress

Table 8: Training progress for transfer learning

| Epoch | Iteration | Time Elapsed (hh:mm:ss) | Mini-batch Accuracy | Mini-batch Loss | Base Learning Rate |
|---|---|---|---|---|---|
| 1 | 1 | 00:00:14 | 17.78% | 1.7788 | 0.0100 |
| 10 | 39 | 00:10:47 | 100.00% | 0.0113 | 0.0100 |

This transfer learning network was tested with APIs given in table 3 which consists of six actions with 100 APIs for each action. The result is shown in figure 19. Even though false positive cases have increased by wrongly classifying 10 walking actions as falling actions and 17 jogging actions as walking, all the falling actions have been correctly recognized. Even though the overall accuracy is 90.2% all the true positive cases (100%) for fall action have been correctly identified.

Fig.19. Confusion matrix of transfer learning network

Table 9: Comparison of classification accuracy of proposed method with other methods on fall action

| Author | Classifier | Fall Dataset used | Accuracy |
|---|---|---|---|
| Nunez-Marcos et al. [7] | CNN | Multiple cameras | Sensitivity : 96.00% Specificity : 99.00% |
| Lu et al. [8] | 3DCNN + LSTM | Multiple cameras | 99.73% |
| Harrou et al. [2] | GLR-SVM | FDD | 96.84% |
| Hussain et al. [9] | KNN | SisFall | 99.80% |
| Kepski et al. [16] | SVM, KNN | URFD | 100% |

| Abobakr et al. [27] | SVM | URFD | 96.00% |
| Our method | SCNN with transfer learning | Multiple camera + FDD + SisFall + URFD combined | 100% |

Table 9 compares the recognition accuracy obtained by different researchers with different methods. It is observed that our system has performed well for fall action recognition and also for other actions even after transfer learning was implemented.

## 6. Conclusion

A vision-based fall detection system using SCNN and transfer learning is proposed in this paper. The action video sequences are pre-processed by performing background subtraction and intensity adjustments together and then edge features of moving object from each frame is extracted and accumulated together to get action pattern image (API). The performance of SCNN has been enhanced and simplified by providing the pre-processed input to transfer learning network. The experiment conducted for recognizing fall action from the videos collected from different data sets gives action recognition rate of 100%. The experiment also exhibits more than 80% recognition accuracy for other actions even after applying transfer learning method. However the training period is reduced in transfer learning. The proposed work can be extended by detecting falls and everyday living activities using live videos.

## The authors have no conflict of Interest

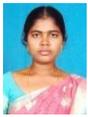

L. Aneesh Euprazia received her B.E. degree in Computer Science and Engineering from St.Xavier's Catholic College of Engineering in 2009 and M.E. degree in Software Engineering from Anna University, Trichy in 2011. She is a full-time Ph.D. research scholar in RMD Engineering College. Her research interests include video-based action recognition, computer vision, object segmentation, machine learning and deep convolutional neural network

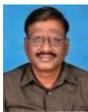

K.K. Thyagharajan received his B.E. degree in Electrical and Electronics Engineering, M.E., degree in Applied Electronics and Ph.D. in Information and Communication Engineering. He is Professor and Dean (Academic) of Electronics and Communication Engineering at RMD Engineering College. He has more than 30 years of experience in teaching, research and administration. He has written 5 books including 'Flash MX 2004' published by McGraw Hill (India). He is reviewer and editorial board member for many International Journals and Conferences. He has published more than 100 papers in National & International Journals and Conferences. He has four patents published to his credit. His research interests include Computer Vision, Semantic Web, Image & Video Processing, Multimedia Streaming, Video Coding, Content-based Information Retrieval, Microcontrollers and e-learning. He is a Fellow of Institution of Engineers (India).